\documentclass[10pt,twocolumn,letterpaper]{article}

\usepackage{wacv}
\usepackage{times}
\usepackage{epsfig}
\usepackage{graphicx}
\usepackage{amsmath}
\usepackage{amssymb}
\usepackage{tabularx}
\usepackage[accsupp]{axessibility}  

\usepackage[T1]{fontenc}
\usepackage{float}
\usepackage{caption}
\usepackage{subcaption}
\usepackage{soul}
\usepackage{booktabs}
\usepackage{mathtools}
\usepackage{appendix}
\usepackage{multirow}
\usepackage[ruled,linesnumbered]{algorithm2e}
\usepackage{pifont}

\usepackage{esvect}
\SetKwInOut{Input}{Input}

\usepackage{bbm}
\usepackage[dvipsnames]{xcolor}

\def\wacvPaperID{745} % Enter the WACV Paper ID here

\wacvfinalcopy % *** Uncomment this line for the final submission

\ifwacvfinal
\fi

\ifwacvfinal
\usepackage[breaklinks=true,bookmarks=false]{hyperref}
\else
\usepackage[pagebackref=true,breaklinks=true,colorlinks,bookmarks=false]{hyperref}
\usepackage{textcomp}
\fi

\definecolor{myred}{RGB}{224, 102, 102}
\newcommand{\method}{T-SINT\,}
\definecolor{custom_green}{HTML}{117733}
\definecolor{custom_red}{HTML}{882255}
\newcommand{\cmark}{{\color{custom_green} \ding{51}}}%
\newcommand{\xmark}{{\color{custom_red} \ding{55}}}%

\newcommand{\myparagraph}[1]{\vspace{0.1cm}{\noindent\textbf{#1}.}}

\begin{document}

%%%%%%%%% TITLE
\title{Learning with Label Noise for Image Retrieval by Selecting Interactions}
\author{Sarah Ibrahimi$^\dagger$\thanks{Work done while interning at NAVER LABS Europe.} \quad Arnaud Sors$^\ddag$ \quad Rafael Sampaio de Rezende$^\ddag$ \quad St\'{e}phane Clinchant$^\ddag$ \\
$\dagger$ University of Amsterdam \quad $\ddag$ NAVER LABS Europe
}

\maketitle
\ifwacvfinal
\thispagestyle{empty}
\fi

%%%%%%%%% ABSTRACT
\begin{abstract}
Learning with noisy labels is an active research area for image classification. However, the effect of noisy labels on image retrieval has been less studied. In this work, we propose a noise-resistant method for image retrieval named Teacher-based Selection of Interactions, \method\footnote{\url{https://europe.naverlabs.com/research/machine-learning-and-optimization/tsint}}, which identifies noisy interactions, \ie elements in the distance matrix, and selects correct positive and negative interactions to be considered in the retrieval loss by using a teacher-based training setup which contributes to the stability. As a result, it consistently outperforms state-of-the-art methods on high noise rates across benchmark datasets with synthetic noise and more realistic noise. 

\end{abstract}

\section{Introduction}
Deep Learning models need large amounts of data to train, but the collection and annotation of large scale datasets is highly time consuming and expensive. One way to avoid data labeling is by employing self-supervised  or semi-supervised learning techniques, that require little to no labels \cite{dino, simsiam, byol}. Another way is to collect annotations from web sources by downloading user-written or automatically generated captions or tags and use them as labels when training supervised models \cite{food101, delf, clothing1m}. These annotations often contain noise due to incorrect captions or tags. As a consequence, noisy labels will prevent models from obtaining their potential best performance. Apart from correcting label noise by manual relabeling, one can think of methods that can distinguish correctly from incorrectly labeled samples during training to avoid the need for manual correction.

\begin{figure}[t!]
\begin{center}
   \includegraphics[width=0.99\linewidth]{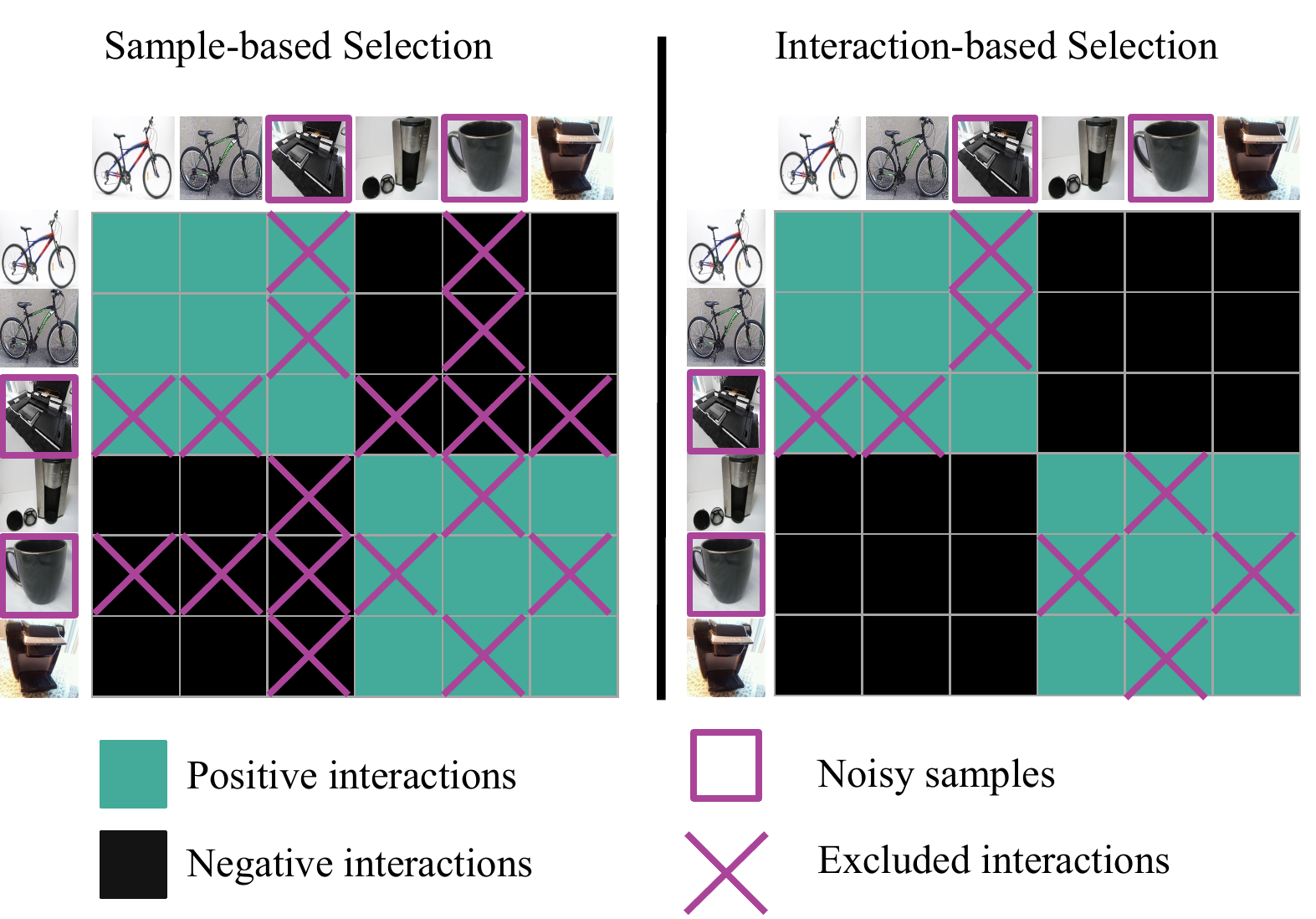}
\end{center}
   \caption{Imagine a set of 6 data samples, with 2 classes (bike and coffee maker) and 3 samples per class. We represent distance matrices between each possible pair in this minibatch and call each pairwise distance an interaction. In this set, one sample of a sofa is mislabeled as a bike and a mug is mislabeled as a coffee maker. Traditional sample-based selection approaches exclude all interactions of these mislabeled samples. Our interaction-based selection approach only excludes false positive interactions, since negative interactions have a high probability to remain negatives after mislabeling. This is indicated by the purple crosses only appearing on the green, positive interactions.} 
   \vspace*{-2mm}
\label{fig:1}
\end{figure} 

Over the past few years, learning with noisy labels has been studied extensively for image classification by using a wide variety of techniques such as robust losses \cite{symmetric, robust}, sample selection \cite{coteaching, mentornet}, regularization \cite{liu2020early,mixup}, and meta learning \cite{learningtolearn}. One of the key ideas of noise-resistant methods for classification is that samples with a high probability of being wrongly labeled should be either ignored or assigned a lower weight during the training process \cite{learning-survey}. The effect of having noisy labels has been less studied for image retrieval. One recent study on noisy labels for image retrieval was done by Liu \etal~\cite{liu2021noise}. Following classification-based methods for learning with noisy labels, it applies the principle of discarding samples with noisy labels on image retrieval and this method performs better than not using any noise-resistant component during training. 

An important difference between classification and retrieval methods in general is that retrieval methods use \textit{interaction losses}, \ie loss functions defined on a pair or tuple of samples, compared to sample-based losses for classification \cite{gordo, musgrave}. Interaction losses aim to bring representations of samples from the same class (positives) closer together and those of different classes (negatives) further apart. Using sample-based selection mechanisms for retrieval has two main disadvantages. First, discarding samples will not only result in removing false positive interactions, but also many true negatives. For large datasets with small noise rates, sufficiently many interactions will remain to learn effectively. However, for high noise rates or small datasets, sample-based techniques might remove too many samples and will most likely break the learning process. The difference between these two approaches is visualized in Fig. \ref{fig:1}. Second, some datasets only contain labels for positive and negative pairs and have no class labels, such as SfM-120K~\cite{radenovic2018fine} for landmark retrieval and MSMARCO~\cite{nguyen2016ms} for document retrieval. Using interactions provides a solution for these two weaknesses by keeping true negative interactions and removing the need for class labels.

In this work we present \textit{\method}, which stands for \textbf{t}eacher-based \textbf{s}election of \textbf{int}eractions for learning with label noise. \textit{\method} selects true positive and negative interactions to be considered in the retrieval loss by using a teacher-based training setup. This simple yet effective mechanism allows us to train robust models even under high noise rates. We show that our method is easy to tune and much more stable than existing methods across different datasets and different noise levels. We evaluate these on several datasets under uniform noise and close to realistic noise. For uniform noise, we beat the best competing method by almost 7\% on average on all noise levels and more than 13\% on the highest noise level. For more realistic noise, we beat the best competing method by up to 3.5\%.

\section{Related work}

\myparagraph{Noisy labels in image classification}
Noise-resistant methods for image classification use a wide variety of techniques such as robust losses \cite{symmetric, robust}, sample selection \cite{understanding-utilizing, coteaching, mentornet,  CurriculumNet, coteaching+}, regularization \cite{hendrycks_pretraining,liu2020early,mixup}, and meta learning \cite{learningtolearn}. All methods aim to detect noisy samples. A common way to do this is by using a small clean subset of the data and learn characteristics about this set \cite{clean1,clean2}. Clean subsets are available in common benchmark datasets for image classification such as Clothing1M \cite{clothing1m}, but in real world scenarios they often don't exist. Another detection technique for noisy samples, is the so called `small-loss trick' \cite{learning-survey}. This trick separates correctly labeled samples from wrongly labeled samples by considering the loss values, since correctly labeled samples have in general smaller loss values. This also relates to the memorization effect \cite{memorization}, stating that Deep Neural Networks learn from clean samples first and then learn from noisy samples.

\myparagraph{Noisy labels in image retrieval}
Learning with noisy labels has not been extensively studied for image retrieval, although two recent works took a first step \cite{superloss, liu2021noise}. The SuperLoss \cite{superloss} is a general noise resistant method that can be used for various tasks such as regression, classification, object detection, and retrieval. It consists of a loss function that has to be applied on top of a task-specific loss function and acts similar to an activation function by reducing the importance of hard, and therefore probably wrongly labeled, samples. For retrieval, the SuperLoss uses an interaction-based selection approach by weighing interactions. While the concept of the SuperLoss is easy and widely applicable, the image retrieval experiments in \cite{superloss} have shown to be highly sensitive to techniques such as hard-negative mining and hyperparameter tuning. Liu \etal~present to the best of our knowledge the first study that solely focuses on noise-resistant image retrieval \cite{liu2021noise}. It shows that existing noise-resistant methods for classification \cite{coteaching,patrini, coteaching+} are not effective for retrieval and are in some cases even worse on image retrieval than commonly used interaction losses without an explicit noise-resistant component. PRISM, the noise-resistant method of \cite{liu2021noise} is a class-center approach that compares the similarity of a feature with all other features of the same class with the help of a memory bank \cite{xbm}, to determine whether the sample is noisy. Any samples detected as noisy are discarded during training.

Where PRISM discards noisy samples and the SuperLoss weighs interactions based on noise estimations, our method combines the best of both approaches. Like the SuperLoss, our method uses an interaction-based selection approach and therefore does not discard true negative interactions as happens with sample-based selection approaches such as PRISM. Our work differs from the SuperLoss by not using a weighing scheme for interactions. By selecting or discarding interactions, like PRISM, we present a more stable method that requires less hyperparameter tuning. Also, our method does not require a memory bank.

\begin{figure*}[ht!]
\begin{center}
   \includegraphics[width=0.99\linewidth]{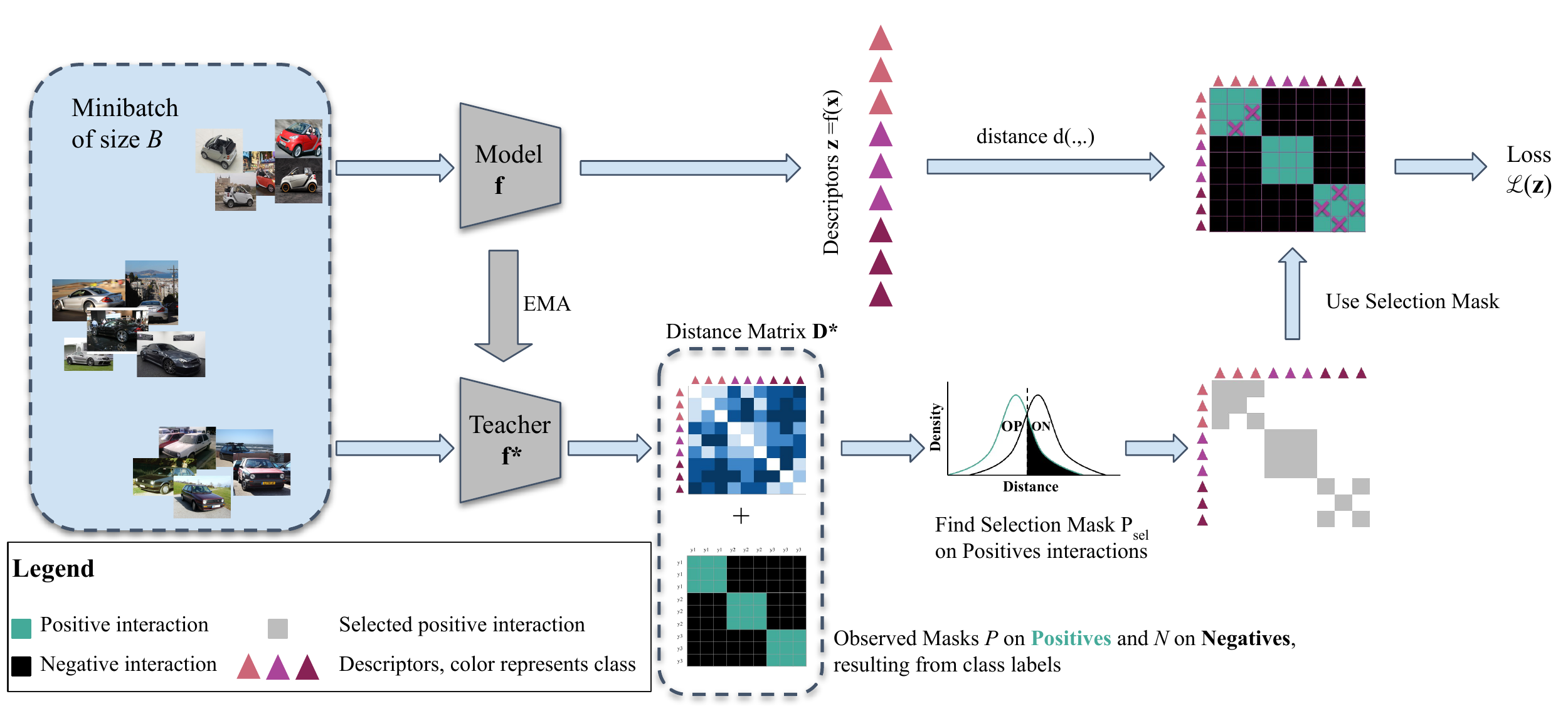}
\end{center} 
   \caption{A model $f$ extracts descriptors that are used in a distance function on all pairs of descriptors to obtain a pairwise distance matrix and to build relevance annotations from class annotations. A teacher model $f^\star$ computes the interaction matrix for the same samples in the batch as the main model. We deduce a selection mask with the help of a cutting value and apply it on the distance matrix of the main model to then let it calculate the loss. Best viewed in color.}
\label{fig:method}
\end{figure*}

\myparagraph{Self-distillation}
Knowledge distillation has gained much popularity across application domains such as computer vision and natural language processing \cite{seed, learning_distillation, s2sd, self-distillation-label-smoothing}. It aims to transfer knowledge or features learned from a teacher network to a student network. When the teacher and student have identical architectures, this is called self-distillation. Fang \etal show how distillation can help in a self-supervised setup \cite{seed}. Li \etal~ present how distillation can be used to learn from noisy labels for image classification \cite{learning_distillation}. With the help of a small set of images with clean labels and a knowledge graph, a model is guided to learn from the entire noisy dataset. Nguyen \etal~ use the model ensembling combined with a prediction ensembling method together with a filtering mechanism for noisy labels to create a noise-resistant method \cite{self}. 

Our method uses a teacher-student setup with a model ensemble technique as presented in \cite{self, mean-teacher}. However, where distillation methods aim for consistency between the teacher and the student output, our method does not need this constraint and only requires a task specific loss.

\section{Method}
\label{sec:method}

Our method starts with a typical image retrieval pipeline. We have a model $f$ that extracts descriptors and we use a distance function on all pairs of descriptors to obtain a pairwise distance matrix and to build relevance annotations from class annotations. This is shown in the top part of Figure \ref{fig:method}. The relevance annotation for a batch is presented by the interaction matrix on the right, where green interactions represent predicted positive pairs and black interactions negative pairs. Since we deal with noisy samples, some of the interactions are incorrect and should be discarded for optimal performance of the model. The key component of our method produces a mask used to select interactions (bottom branch of Figure \ref{fig:method}), which is then applied to interactions in the main branch. This is further described later in this section. 

\subsection{Loss}
\label{sec:retrieval_pipeline}

Our method is based on the contrastive margin loss \cite{ctrm}, except it calculates it on \textit{selected} interactions rather than using them all. Before we explain how to select,  we introduce this loss. 
Image retrieval methods use interaction losses that bring descriptors of samples from the same class closer together and those of different classes further apart.
For a minibatch $\mathcal{B}=\{(x_1,y_1),...,(x_B,y_B)\}$ of size $B$, we 
can compute the real-vector feature representations $z_i=f_\theta(x_i)$ at the output of model $f$ parameterized by $\theta$. 
We denote $\mathbf D$ the $B\times B$ matrix of all possible pairwise distances between elements of $\mathcal{B}$:
\begin{equation}
    \mathbf D = \{D_{i,j}, (i,j) \in [\![1, B]\!]^2 \ | \ D_{i,j}=d(z_i, z_j)\},
    \end{equation}
where $d$ is usually the euclidean or cosine distance.

For any batch indices $1 \leq i, j \leq B$, the contrastive margin loss works on all pairs of descriptors $(z_i, z_j)$ as
\begin{align}
    \begin{cases}
	\ell_{i,j} &= 1\! \! 1_{y_i=y_j}  \ell_p(z_i, z_j) + 1\! \! 1_{y_i \neq y_j} \ell_n(z_i, z_j); \\
	\ell_p(z_i, z_j) &= {D_{i,j}}^q; \\
	\ell_n(z_i, z_j) &= \max(0, m - D_{i,j})^q,
	\end{cases}
\end{align}
where the form of positive terms ($\ell_p$) and negative terms ($\ell_n$) are different,
$m>0$ is the margin and the distance exponent $q$ is usually $1$ or $2$ (in our case $1$). Denoting by $\mathcal P=\{(i,j) \in [\![1,b]\!]^2 | y_i=y_j \}$ the set of positive pairs and
$\mathcal N= \{ (i,j) \in [\![1,b]\!]^2 | y_i \neq y_j\}$ the set of negative pairs, 
the loss $\mathcal{L}$ over the batch is an aggregation of positive and negative interactions:
\begin{align}
    \begin{cases}
    \mathcal L(\mathbf z) &= B^{-2}(\bar\ell_p(\mathbf z) + \bar\ell_n(\mathbf z)); \\
    \bar\ell_p(\mathbf z) &= \mathbb{E}_{(i,j)\sim\mathcal{U}(\mathcal P)}[\ell_p(z_i, z_j)]; \\
    \bar\ell_n(\mathbf z) &= \mathbb{E}_{(i,j)\sim\mathcal{U}(\mathcal N)}[\ell_n(z_i, z_j)].
    \label{eq:contrastive_means}
    \end{cases}
\end{align}

\subsection{Interaction Selection}
\label{sec:is}
When a dataset has noisy labels, it will affect interactions. For a sample with a corrupted label, when the number of classes is large an observed positive interaction using this sample is almost always a false positive. For negative interactions, this has less impact: whenever the number of different instances in a batch is much smaller than the number of total classes in the dataset, chances are very small that an observed negative interaction is actually a positive interaction. A study on false negative interactions is provided in the Supplementary material.

We are working on a distance matrix $\mathbf D$ where \textit{observed positives} elements $\{D_{i,j} \ | \ (i,j) \in \mathcal P\}$ are noisy and observed negative elements $\{D_{i,j} \ | \ (i,j) \in \mathcal N\}$ can be considered all clean.
We would like to identify \textit{false positive} interactions and exclude them from the aggregation in the loss function.
As many noise correction methods \cite{superloss, coteaching, mentornet, coteaching+}, we rely on the fact that if we have a trained retrieval model $f^\star$, the \textit{distance value} on positive elements
according to this model gives us an indication on the likelihood of the interaction being a true or a corrupted one.
Interactions between clean samples are expected to have a small distance value and for interactions that are noisy the distance values will most likely be larger.

\begin{figure}[t!]
\begin{center}
   \includegraphics[width=0.99\linewidth]{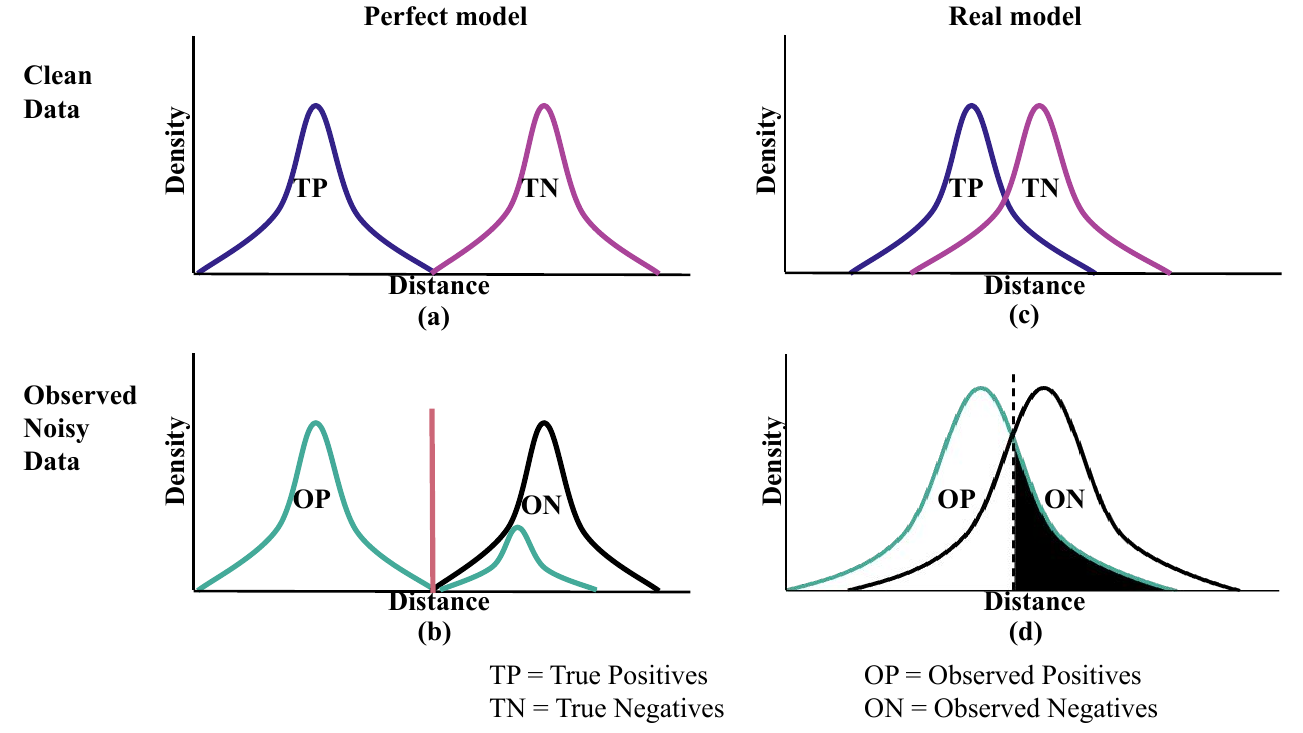}
\end{center}
   \caption{A perfect retrieval model, represented on the left, is also able to identify noise. A real model can approach this idea. Section \ref{sec:is} details this.}
\label{fig:distributions}
\end{figure}

Figure~\ref{fig:distributions} exemplifies this idea. (a) Consider two distributions $p_{TP}(d)$ and $p_{TN}(d)$ of true positive and true negative interactions respectively. If we have at our disposal a hypothetical \textit{perfect} model in the sense of a retrieval metric, it will attribute smaller distance values to all true positive interactions compared to any negative interaction. (b) Therefore, when in an observed setting, we can find a distance value of this model that can be used to separate true positive from false positive interactions (amongst observed positives), as indicated by the red line. (c) In actual fact, such a perfect model is obviously not available, but we can still use a non-perfect $f^\star$, according to which $p_{TP}(d)$ and $p_{TN}(d)$ will likely overlap for difficult interactions,
but which can already help identify easier noisy interactions. (d) By using a cutting value, a subset of the positive interactions will be removed and therefore not considered by $f$.

\subsection{Using a Teacher}

Now comes the question of what to use for $f^\star$. In practice a perfect model $f^\star$ is not available, otherwise the retrieval problem would already be solved.
To create a model which approaches it, we take inspiration from knowledge distillation, where a strong teacher network provides knowledge to a student network that learns the knowledge \cite{distillation_hinton}. More specifically, we take inspiration from Mean Teacher \cite{mean-teacher}. It consists of a teacher-student setup with two similar model architectures, where the student model weights are updated through gradient descent and the teacher model weights as an exponential moving average of the student weights. Mean Teacher uses two loss functions, a classification loss on the student model and a consistency loss that compares the softmax outputs of both networks, usually the MSE loss or KL-divergence loss. Strictly speaking, Mean Teacher is not a self-distillation method, since it optimizes for consistency during training, while for self-distillation methods the consistency regularization is usually performed after training. 

We decide to use a teacher only for the task-specific loss and do not aim to optimize for consistency between the teacher and student models outputs. Therefore, we do not consider this as a typical knowledge distillation approach. Our main model and teacher model have the same architecture and same initialization. Our main model is updated by backpropagation and we choose the update for the teacher model to be an exponential moving average of the parameters of the main model every iteration.
\begin{equation}
\theta^\star_{t} = \alpha\theta^\star_{t-1} +(1-\alpha)\theta_{t},
\end{equation}
where $\theta, \theta^\star$ are the main model and teacher's weights, respectively. Our network updating step is similar to~\cite{mean-teacher}.

\subsection{Selection algorithm}
\label{sec:is}

\begin{algorithm}[t!]
\caption{Teacher-based selection of interactions: calculation of the minibatch loss}\label{alg:nsm}
	\SetAlgoLined
    \SetKwInOut{Input}{input}\SetKwInOut{Output}{output}
    \SetKwProg{Init}{initialization}{}{}
	\Input{$\mathbf z = \{z_i=f(x_i) \, \ i \in [\![1, B]\!]\}$, $\mathbf z^\star = \{z_i=f^\star(x_i) \, \ i \in [\![1, B]\!]\}$, $\mathbf y = \{y_i, i \in [\![1, B]\!]\}$: a minibatch of main model features, teacher model features, and class labels;\\
     $\tau$: selection hyperparameter;\\$\beta$: momentum on cutting threshold upd.;\\$d_{cut}$: current value of cutting threshold;\\
     $\mathbf P, \ \mathbf N \in \{0,1\}^{B\times B}$ : observed masks;
     }
    \Output{loss value $\mathcal L(\mathbf z)$}
    \Init{$d_{cut} \leftarrow$ None} 
    
    Build distance matrices between all pairs of features, for both the main model and teacher model:
    $\mathbf D = \{D_{i,j} \ | \ i, j \in [\![1, B]\!]^2, \  D_{i, j} = d(z_i, z_j)\}$
    $\mathbf D^\star = \{D^\star_{i,j} \ |\ i, j \in [\![1, B]\!]^2, \  D^\star_{i, j} = d(z^\star_i, z^\star_j)\}$ \\

    Gather observed positive distance values according to \textbf{teacher} model:
    $\mathcal D^\star_P = \{D^\star_{i,j} \ | \ P_{i,j}=1\}$\\
    Denote $F$ the cumulative distribution of $\mathcal D^\star_P$, calculate the $\tau$-percentile of $F$: 
    $d_B=F^{-1}(\tau)$
    \\
    Update cutting value with moving average: \quad \quad \quad  \textbf{if} $d_{cut}$ is None \textbf{then} $d_{cut} \leftarrow d_B$; \textbf{else} 
    $d_{cut} \leftarrow \beta d_{cut} + (1-\beta) d_B$ \\
    Deduce selection mask on positives: $\mathbf P_{sel}=\mathbbm{1}(\mathbf D^\star< d_{cut}) \ \& \ \mathbf P$\\
    Calculate loss using the \textbf{main model's} distance matrix:
    \begin{align*}
    &\bar\ell_p(\mathbf z) = \mathbb{E}_{\mathbf P_{sel}=1}[D_{i,j}]; \\
    &\bar\ell_n(\mathbf z) = \mathbb{E}_{\mathbf N=1}[\max(0, m - D_{i,j})]; \\
    &\mathcal L(\mathbf z) = B^{-2} (\bar\ell_p(\mathbf z) + \bar\ell_n(\mathbf z))
    \end{align*}
\end{algorithm}

As described in Section~\ref{sec:retrieval_pipeline} we want to select and keep only a subset of interactions amongst \textit{observed positives}, according
to their distance values in the output space of a teacher model $f^\star$.
To do so, we could simply select out the top-$K$ interactions in terms of distance value, with $K$ chosen such as to reach a predefined selection ratio. However, if we have a dataset with noise rate $r$, because minibatches have a limited size, each of them will not contain exactly $rB$ noisy samples but a number which varies around $rB$ from minibatch to minibatch.
In order to allow for a selection ratio which also varies from minibatch to minibatch, if we want to keep on average a proportion $\tau$ of positive interactions, we keep a multiplicative running-average $d_{cut}$ (with momentum $\beta$) of the $\tau$-percentile of observed positive distance values, and do the selection on each minibatch using this cutting value.
Algorithm~\ref{alg:nsm} details this selection step.

Now we explain how to choose the threshold $\tau$. In general, noise-resistant methods that use the loss value as a selection mechanism to find noisy samples, use one or more hyperparameters to set the desired selection level. This is necessary to decide which samples to ignore or how to weigh them within a batch. When applying the method to a practical problem, it is useful to have guidelines about how to choose these hyperparameters. For example, \cite{liu2021noise} uses the noise rate that corresponds to the noise level they introduced in their synthetic uniform noise.
In our approach, we make the following estimation. Assume we have a noise rate $r$ on examples in our dataset and $k$ instances per class in each mini-batch. Then we can estimate the proportion of true positives amongst all observed off-diagonal positives as $\tilde p \sim (1-r)^{2},$ or equivalently including the diagonal,
\begin{equation}
\tilde p \sim \frac{(1-r)^{2} (k^2-k) + k}{k^{2}}, 
\label{eq:ptp}
\end{equation}
where $k^{2}$ is the total number of interactions for an instance, $(1-r)^{2}$ is the estimated proportion of clean interactions and the $(k^{2}-k)$ multiplier and the $1/k$ addition account for the fact that noise affects only non-diagonal elements, although the selection is performed on all. This expression is valid if the per-batch noise level has low variance around $r$, i.e. when $B>>1$, and when the number of classes is sufficiently high so that the case where two samples forming a positive interaction get corrupted to the same class is rare.

$\tau=\tilde p$ constitutes a reasonable starting value for the selection hyperparameter if an estimate of the noise rate on class annotations is available.
However, this is only an initial guess, and the best value may vary a little.
Indeed, the above calculation pre-supposes that the optimal number of interactions to filter is equal to the true number of corrupted interactions.
This is true in the case of a perfect teacher, but in the case of a real teacher this may not exactly be.
Also, the noise rate in various experimental datasets may be slightly different than the expected one.

\subsection{Comparison with state of the art}
\label{sec:comparison}

We compare \method with two existing noise-resistant approaches for image retrieval: SuperLoss and PRISM.
\myparagraph{SuperLoss~\cite{superloss}} This is a loss that can be applied on top of any other loss. Assume we have an input loss value $l_{i}$, then the SuperLoss SL can be applied in the following way

\begin{equation*}
    SL_{\lambda}(l_{i}) = \min_{\sigma_{i}} \left((l_{i} - \tau)\sigma_{i} + \lambda(\log \sigma_{i})^{2}\right),
\end{equation*}

where $\tau$ is a threshold that separates easy from hard samples based on their loss values, $\lambda$ is a regularization hyperparameter and $\sigma_{i}$ is a confidence parameter which in practice will be replaced by the converged value at the limit \footnote{An analytical expression is given in \cite{superloss}.}. Therefore only $\tau$ and $\lambda$ should be tuned, where $\tau$ is usually a running average of the input loss during training.
 
Whenever the SuperLoss is applied on an interaction loss, such as the contrastive loss, this can be described as follows:
 
  \resizebox{.98\hsize}{!}{$SL_{\lambda}^{CL}(f(x_{i}),f(x_{j}),y_{ij}) = \begin{cases}SL_{\lambda}(l_{+}^{CL}(f(x_{i}),f(x_{j})))  \text{ if }y_{ij}=1, \\ 
     SL_{\lambda}(l_{-}^{CL}(f(x_{i}),f(x_{j}))) \text{ if }  y_{ij}=0.\end{cases}$}
where the two losses use two independent thresholds $\tau_{+}$ and $\tau_{-}$ and share the same $\lambda$ value.

\myparagraph{PRISM~\cite{liu2021noise}} This method starts with estimating which samples are clean. Therefore it compares each sample representation $z_{i}$ with all stored features $v_{j}$ of the same class $k$ in the memory bank $M$. 
When the estimated probability, calculated by the Softmax of a sample being clean, $P_{\text{clean}}(i)$, is below a threshold $m$, it will be considered as a noisy sample and discarded. $m$ is calculated through a smooth top-R method that uses an estimated noise rate $R$. Specifically,

\begin{equation*}
m=\frac{1}{\tau}\sum_{j=t-\tau}^{t} Q_{j},
\end{equation*}
where $Q_{j}$ is the $R^\text{th}$ percentile $P_{\text{clean}}(i)$ value in the $j$-th minibatch and $\tau$ stands for the number of last batches. 

After selecting the clean data samples according to the threshold $m$, these are used to calculate the selected loss, for example the contrastive loss. Furthermore, a memory bank loss is calculated, to enable more positive and negative pairs in the loss. The main hyperparameter to tune or estimate is $R$, which is compared to the SuperLoss more intuitive. 

\section{Experiments}

\subsection{Datasets}
We report results on four commonly-used datasets for image retrieval \textit{Caltech-UCSD Birds-200-2011 (CUB)} \cite{cub}, \textit{CARS} \cite{cars}, \textit{Stanford Online Products (SOP)} \cite{sop}, \textit{Revisited Paris (RParis6k)} \cite{rparis6k} and a recent dataset created for learning with label noise for image retrieval, \textit{CARS-98N} \cite{liu2021noise}.

CUB consists of 11,788 images in 200 classes, where the first 100 classes are used for training and the rest for evalation. CARS has 16,158 images with 196 classes, where the first 98 are used for training and the remaining classes for evaluation. SOP contains 120,053 product images from 22,634 classes. The first 11,318 classes (59,551 images) are for training and the other 11,316 (60,502 images) classes are used for evaluation. 

RParis6k consists of  6,412 images of 12 landmarks in Paris. This dataset is a cleaned version of the Paris dataset~\cite{paris} provided by Radenovi\'{c} \etal~\cite{rparis6k}. We use this dataset for evaluation when training on two noisy datasets with landmarks: Oxford \cite{oxford} and Landmarks \cite{landmarks}. The Oxford dataset consists of 5,062 images of 11 Oxford landmarks. We use the original version of this dataset that has not been cleaned.  The Landmarks dataset, also known as Landmarks-full, consists of 167.231 images in its train set. These images are retrieved by querying a search engine and therefore considered as noisy.

CARS-98N is a recent training set with noisy labels, created by crawling 9,558 images for 98 car models from Pinterest. These models correspond to the 98 labels from the CARS training set. Since the test set of the CARS dataset is considered as clean, it is used for evaluation.

\begin{figure*}[ht!]
\begin{center}
   \includegraphics[width=0.90\linewidth]{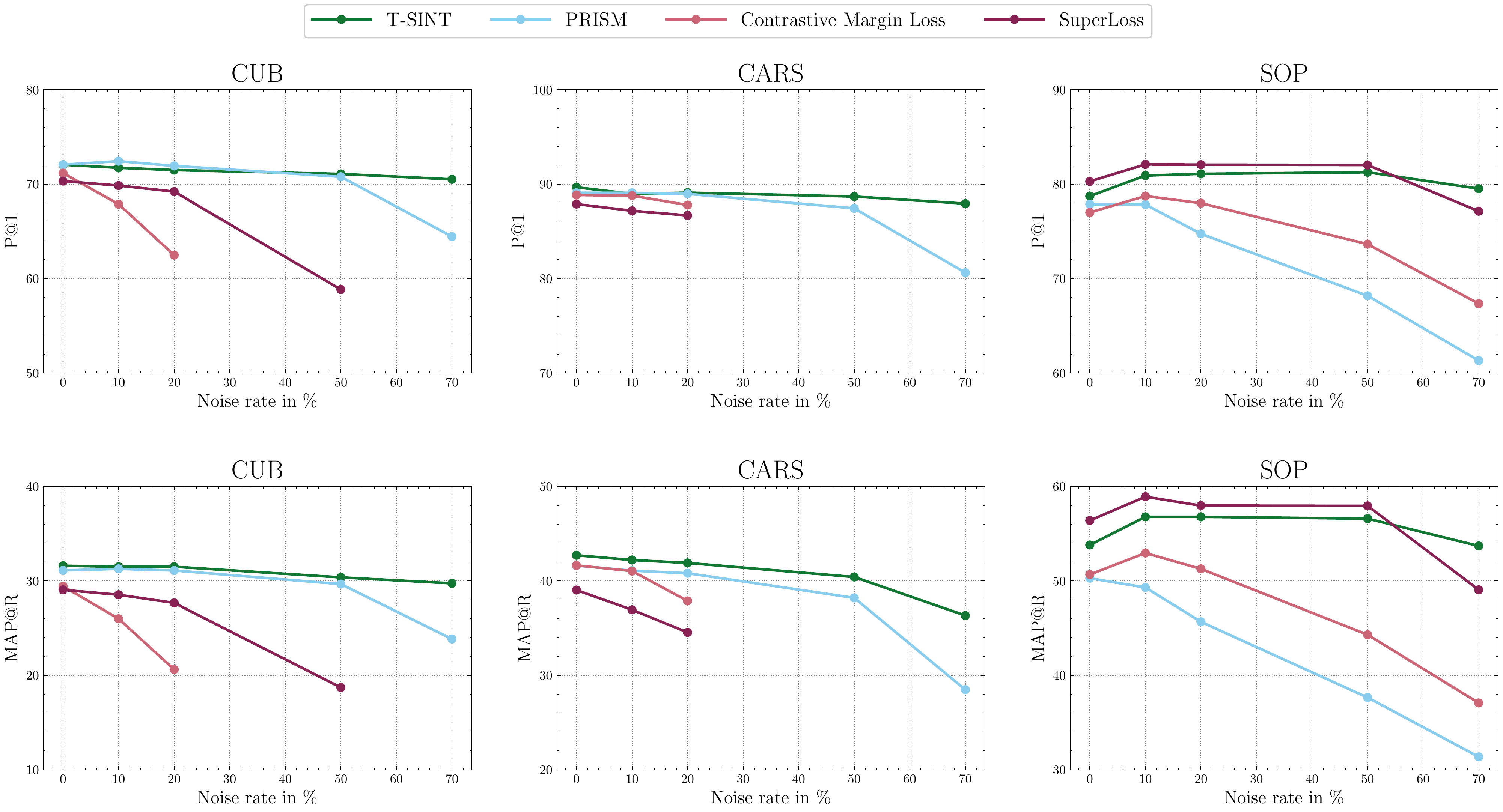}
\end{center}
   \caption{\textbf{Precision@1} ((a),(c),(e)) and \textbf{MAP@R} ((b),(d),(f)) results on CUB, CARS and SOP respectively. 
   All methods depicted use the CLIP model as a teacher.
   Whenever a method has no measure point for a specific noise rate, it means that the model was not able to perform better than its initial state when using the CLIP model. This is the case for the Contrastive Margin Loss and the SuperLoss on CUB and CARS. Note that the y-axis has a different range for each dataset and metric. Best viewed in color.}
\label{fig:uniform_results}
\end{figure*}

\subsection{Noise Types and Rates}
\label{sec:noise_types}
CUB, CARS, and SOP are considered as clean datasets and have only been used in a noise study by \cite{liu2021noise} by adding synthetic noise to the train sets of these datasets. However, we do not rule out that any noise in these original datasets is present. We follow \cite{liu2021noise} by applying 10\%, 20\%, and 50\% uniform noise to these datasets. Following noise studies for classification, we add 70\% of uniform noise to our study. 

More realistic noise is present in Oxford, Landmarks and CARS-98N, since their images are collected with the help of search engines and not manually cleaned. Radenovi\'{c} \etal showed that the Oxford dataset has noise in its annotation that is not related to the building category but to the difficulty of recognizing the building which is indicated by 'positive', 'junk', or 'negative' \cite{rparis6k}. There is no exact estimation of the noise rate. Gordo \etal~\cite{gordo} showed that Landmarks contains a non-negligible amount of unrelated images. A clean version of this dataset was presented by \cite{gordo} and left only 25\% of the training images and 87\% of the classes. We use this as a guideline to estimate the noise level in this dataset. For evaluation purposes we use the cleaned test set of the Paris dataset \cite{paris}, RParis6k, that has been cleaned manually by \cite{rparis6k}. The level of noise in CARS-98N has been estimated by the creators of this dataset at 50\%.

\subsection{Baselines \& Implementation Details}

We compare our method \method against three methods:
\begin{itemize}
    \item PRISM \cite{liu2021noise}, which is the current State-of-the-Art approach for learning with label noise for image retrieval on the benchmarks we consider.
    \vspace{-1mm}
    \item SuperLoss \cite{superloss}, which has been effective for label noise on image retrieval, in particular on landmarks.
    \vspace{-1mm}
    \item Contrastive Margin Loss \cite{ctrm}, which is a simple baseline without a noise-resistant component. We use this to compare to PRISM, the SuperLoss and \method since the three methods all use the contrastive margin loss.
\end{itemize}

\method uses a CLIP \cite{clip} model with a ViT-B/32 backbone \cite{vit} and a similar teacher, both without a head, resulting in a 512 dim output feature. We choose CLIP since it covers a wide range of domains and are therefore more suitable than domain specific image retrieval datasets. We rerun all baseline methods by using the same CLIP model as a backbone. Results on the original backbones as described in \cite{liu2021noise} can be found in the Supplementary material. 

For all methods, we tune the learning rate and the batch size. Details about the best values of these hyperparameters for each method and dataset are provided in the Supplementary material. The noise-resistant methods need some additional tuning. As explained in Section \ref{sec:comparison}, the SuperLoss needs tuning for the regularization parameter $\lambda$ and two thresholds $\tau_{+}$ and $\tau_{-}$ for the positive and negative interactions. For each dataset and each noise level, we did a hyperparameter search for $\lambda$ and $\tau_{+}$ and $\tau_{-}$. For $\lambda$ we tried the values 0.001, 0.01, 0.05, 0.1, 0.25, 1.0 as these values were recommended in \cite{superloss}. For the thresholds, \cite{superloss} recommends three options: a global average, an exponential running average with a fixed smoothing parameter or a fixed value given by prior knowledge on the task.  Therefore, we experimented with the global average and the exponential running average. The best values can be found in the Supplementary material. For PRISM, we use the estimated noise rate as provided \cite{liu2021noise}. For Oxford and Landmarks we tune this value. For \method, we estimate the proportion of true positives by equation \ref{eq:ptp}.

Following \cite{liu2021noise, musgrave}, we use Precision@1 (P@1), Mean Average Precision@R (MAP@R), and mean Average Precision (mAP) for evaluation, since a ranked list of nearest neighbours on the test set is used for evaluation.

\subsection{Comparison with State-of-the-Art}

\myparagraph{Uniform Noise} We analyze the effect of several levels of uniform noise (0\%, 10\%, 20\%, 50\%, and 70\%) on CUB, CARS, and SOP. The results are presented in Figure \ref{fig:uniform_results}. Note that the scores for 0\% noise are reference scores that we aim to match in the case of label noise.

In general, the CLIP model performs quite well on all three datasets. However, the CLIP model is not noise-resistant. Combined with the contrastive margin loss, it performs well on 0-10\% noise, but starts to drop severely for larger noise rates. For CUB and CARS, this method is not even able to achieve a score higher than when using a CLIP model as a feature extractor, which is indicated in Figure \ref{fig:uniform_results} by missing points. Therefore we can conclude that a strong backbone is not sufficient by itself to be resistant to noise. 

Considering PRISM, we observe that for CUB and CARS, \method performs on par with this method on noise levels of 10\%, 20\% and 50\%. However, for 70\% noise, PRISM shows a large drop in performance compared to 50\% noise, whereas our method is much more stable. The gap between PRISM and our method is 5 to 6 points for both the P@1 and the MAP@R scores. For SOP we notice a larger gap over all noise levels, where in the case of 70\% noise our method outperforms PRISM by more than 20\% for P@1 and MAP@R scores. On average on CUB, CARS, and SOP at 70\% noise, our method scores more than 10\% higher on P@1 and more than 13\% higher on MAP@R.

\begin{table}[t]
\centering
\resizebox{\columnwidth}{!}{%
\begin{tabular}{@{}lcrrr@{}}\toprule
Method & Training set & Easy & Medium & Hard \\ \midrule
\multirow{2}{*}{$\text{CTRM}_{\text{BN-inception}}$\cite{ctrm}}      & Oxford & 64.66 & 50.37 & 25.26\\ 
& Landmarks & 77.33 & 62.67 & 37.94\\
\hline
\multirow{2}{*}{$\text{PRISM}_{\text{BN-inception}}$ \cite{liu2021noise}} & Oxford & 62.72 & 49.28 & 24.38 \\ 
& Landmarks & 76.33 & 62.03 & 37.05 \\
\hline
\multirow{2}{*}{$\text{CTRM}_{\text{ViT-B/32}}$ \cite{ctrm}} & Oxford & 74.90 & 62.63 & 38.40 \\
& Landmarks & 85.62 & 73.53 & 51.27 \\
\hline
\multirow{2}{*}{$\text{PRISM}_{\text{ViT-B/32}}$ \cite{liu2021noise}} & Oxford & 77.92 & 65.50 & 41.74 \\
& Landmarks & 85.52 & 72.99 & 49.52 \\
\hline
\multirow{2}{*}{$\text{SuperLoss}_{\text{ViT-B/32}}$ \cite{superloss}} & Oxford & \textbf{82.59} & \textbf{70.35} & \textbf{47.68} \\
& Landmarks & 86.72 & 76.27 & 56.11\\
\hline
\multirow{2}{*}{\method$_{\text{ViT-B/32}}$~\text{(Ours)}} & Oxford & 82.21 & 70.18 & 47.36  \\
& Landmarks & \textbf{87.55} & \textbf{77.66} & \textbf{57.17}\\
\bottomrule
\end{tabular}
}
\caption{mAP scores for training on Oxford and Landmarks, testing on RParis-6k. 
}\label{tab:oxford}
\vspace{-1mm}
\end{table}

The SuperLoss performs significantly worse than our method on CUB and CARS for 10\% and 20\% noise. For CUB the gap even increases to a difference of 10\% in P@1 and MAP@R at 50\% noise. However, for 70\% uniform noise on CUB and for 50\% and 70\% uniform noise on CARS using the SuperLoss breaks the training at each combination of hyperparameters that has been tested. We have no clear indication why the set of hyperparameters that we use and has been suggested by \cite{superloss} for the SuperLoss does not work on these noise rates for CUB and CARS. However, Castells \etal~\cite{superloss} indicate that some learning setups for image retrieval for this method lead to poor performance and suggest to therefore tune more elements in the training setup including explicit hard negative mining, GeM pooling and descriptor whitening. We decided not to add these components and additional tuning for the SuperLoss to be able to make a fair comparison to our method and PRISM. This additional tuning is not necessarily required for all datasets. For example on SOP, we see that training with the SuperLoss does not break the training and even slightly outperforms our method, except for a noise rate of 70 \%. Nevertheless, our method T-SINT is the only method that does not show a significant drop in performance from 0\% to 70\% noise for SOP and is therefore the most stable.

\myparagraph{More realistic noise} We study the effectiveness of our method on more realistic noise by experiments on Oxford, Landmarks, and CARS-98N. For experiments on Oxford and Landmarks we evaluate on RParis6k by differentiating between easy, medium and hard samples following \cite{rparis6k}. In Table \ref{tab:oxford}, we see that when training on Oxford, the SuperLoss performs slightly better than our method. However, when training on a much larger dataset, our method outperforms the SuperLoss. In both cases, we see a large gap between PRISM and our method.  

\begin{table}[t]
\centering
\resizebox{0.65\columnwidth}{!}{%
\begin{tabular}{@{}lrr@{}}\toprule
Method                                         & P@1   & MAP@R \\ \midrule
$\text{CTRM}_{\text{BN-inception}}$            & 44.91 & 4.76 \\ 
$\text{PRISM}_{\text{BN-inception}}$ \cite{liu2021noise} & 57.95 & 8.04 \\ 
$\text{CTRM}_{\text{ViT-B/32}}$ & 82.59 & 30.38  \\
$\text{PRISM}_{\text{ViT-B/32}}$ \cite{liu2021noise} & 81.75 & 29.93  \\
$\text{SuperLoss}_{\text{ViT-B/32}}$ \cite{superloss} & 82.55 & 31.31 \\
$\text{T-SINT}_{\text{ViT-B/32}}$~\text{(Ours)} & \textbf{86.10} & \textbf{34.93}  \\ \bottomrule
\end{tabular}
}
\caption{Precision@1 (\%) and MAP@R (\%) for CARS-98N. 
}\label{tab:carsn}
\end{table}

\begin{table}[t]
\centering
\resizebox{\columnwidth}{!}{%
\begin{tabular}{@{}ccccrr@{}}\toprule
\begin{tabular}[c]{@{}c@{}} EMA \\ teacher \end{tabular} & EMA $d_{cut}$ & \begin{tabular}[c]{@{}c@{}} Teacher \\ model \end{tabular}  & \begin{tabular}[c]{@{}c@{}} Model \\ backbone \end{tabular} & P@1 & MAP@R \\ 
\midrule
\cmark & \cmark & ViT-B/32 & ViT-B/32 & 86.10 & 34.39 \\
\xmark  & \cmark & ViT-B/32 & ViT-B/32 & 84.80 & 33.22 \\
\cmark & \xmark & ViT-B/32 & ViT-B/32 & 84.87 & 32.54  \\
\xmark  & \cmark & ViT-B/32 & BN-Inc & 38.93 & 3.98 \\
\cmark  & \cmark & BN-Inc & BN-Inc & 38.10 & 3.60 \\ \bottomrule
\end{tabular}
}
\caption{Ablation study on \method, evaluated on CARS-98N. Precision@1 (\%) and MAP@R (\%).} 
\label{tab:ablation}
\end{table}
Table \ref{tab:carsn} presents the results on CARS-98N. Our method outperforms all other methods by a large margin. Compared to PRISM, we report a gap of almost 5\% for both the P@1 and MAP@R and for the SuperLoss this gap is 3.5\% for both metrics. 

\subsection{Ablation Study}

We present an ablation study on our method to see which components contribute to its performance. The results are presented in Table \ref{tab:ablation}. We notice that updating a teacher by taking the exponential moving average of the parameters in the main model compared to freezing the teacher gains slightly more than 1\% of performance. Updating the cutting value with a moving average compared to keeping the cutting value fixed adds almost 2\% to the score. We also study the effect of backbones. In case our main model and teacher do not have the same backbone architecture, there is no exponential moving average for the teacher's weights. Whenever we keep the ViT-B/32 backbone of the teacher and take a BN-inception pretrained on ImageNet for the main model, our model completely breaks and performs even worse than only using the contrastive margin loss with a BN-inception backbone. The reason for this is unclear so far as T-SINT is in principle compatible with the use of different architectures for the teacher and main model. Whenever we replace both ViT-B/32 backbones of the main model and the teacher by a BN-inception backbone, our method also breaks. This emphasizes that our method requires a good teacher.

\section{Conclusion}
We propose \method, an effective noise-resistant method for image retrieval. With the help of a teacher, it identifies noisy interactions and selects correct positive and negative interactions in the distance matrix to be used by the retrieval loss. Our simple selection mechanism achieves state-of-the-art results on both synthetic noise and more realistic noise,  consistently outperforming existing methods on high noise rates and being the most stable across all noise rates.

{\small
\bibliographystyle{ieee_fullname}
\bibliography{egbib}
}

\clearpage

\ifwacvfinal
\thispagestyle{empty}
\fi
\appendix

\setcounter{table}{0}
\setcounter{figure}{0}
\renewcommand{\thetable}{T\arabic{table}}
\renewcommand{\thefigure}{F\arabic{figure}}

\section{Sensitivity of the $\tau$ hyperparameter}

\begin{figure}[t!]
\begin{center}
   \includegraphics[width=0.99\linewidth]{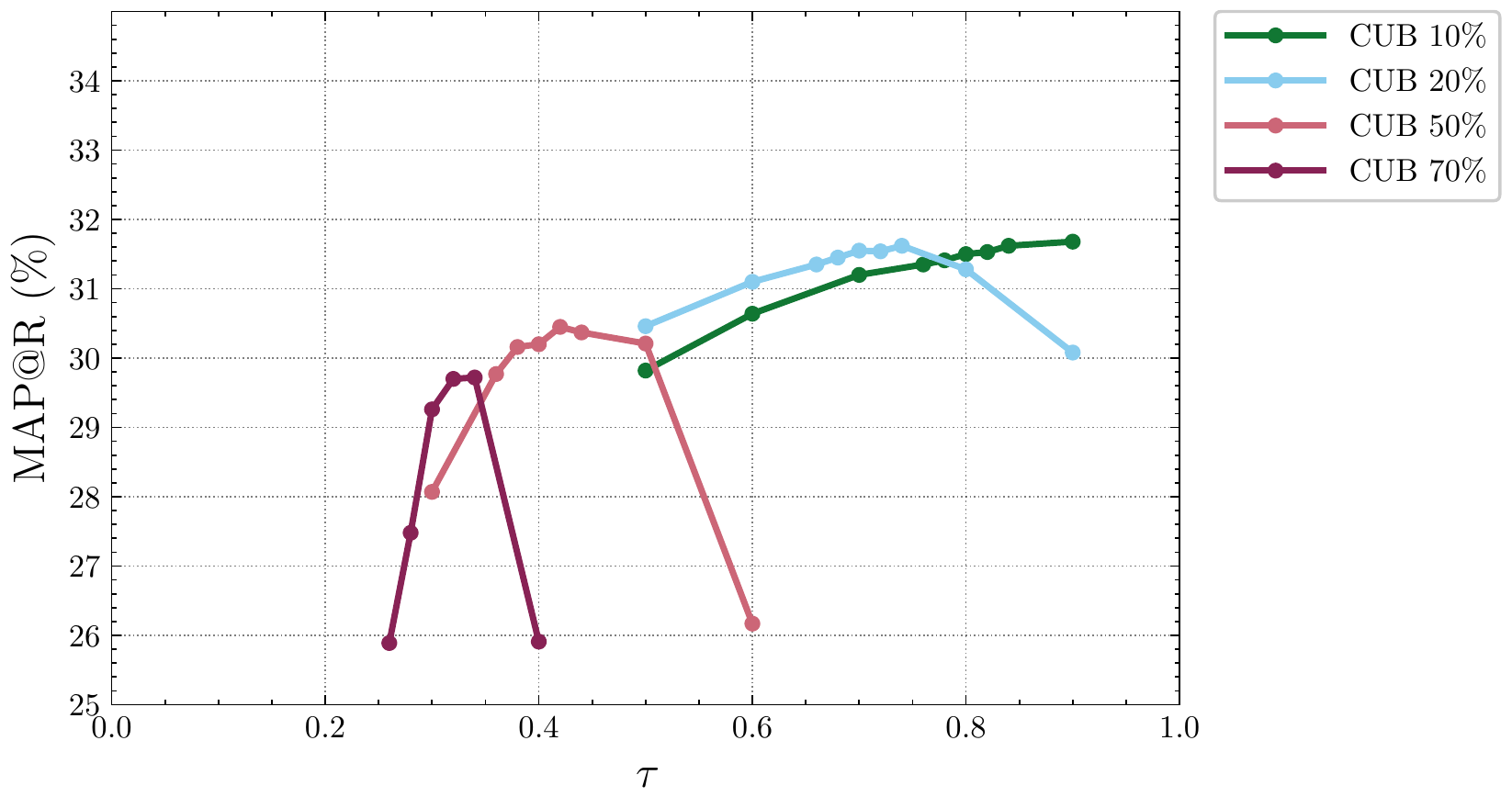}
\end{center}
   \caption{For low noise rates, T-SINT is not very sensitive to the $\tau$ value. For higher noise rates it is more sensitive, but even for 50\% noise it has a range of possible $\tau$ values where the performance is quite stable.}
\label{fig:sensitivity}
\end{figure}

For T-SINT, $\tau$ is tuned, starting with an estimation of its value. We analyze how sensitive T-SINT is to the selection of the $\tau$ for different noise values for CUB by analyzing MAP@R results. This is shown in Figure \ref{fig:sensitivity}.

\section{Negative interactions}
We stated that chances that an observed negative interaction is actually a positive interaction are very small whenever the number of different instances in a batch is much smaller than the number of total classes in the dataset. For T-SINT, we use a batch size of 80 samples, with 4 samples per class, resulting in 20 classes per batch. 

We analyse what is the minimum number of classes for which this assumption holds and we performed experiments on subsets of the CARS train set, consisting of 20, 40, 60, and 80 classes. To each of these subsets, we add the noise ratios as for the full CARS dataset with 98 classes. Fewer classes leads to fewer training images which will result in a drop in performance. Therefore we do not compare the performances between those subsets, but the ratios of performance with noise to performance without noise. This results in Table \ref{tab:negatives}, which shows the Precision@1 scores. From this table, we can see that only for the subset of 20 classes, the relative performance is much lower for increasing noise ratios, which indicates that the method does not work very well in this case. This might be due to using false negative interactions, but might also be caused by the small number of clean samples that are present in the subset of 20 classes for high noise rates. 

From this table, we can conclude, that our method seems to work well for datasets with at least 40 classes, which is common for most real-world image retrieval datasets. For datasets with a smaller number of classes, one could think of adding an additional hyperparameter equivalent to $\tau$ to serve as a threshold for negative interactions. 

\begin{table}[hbt!]
\centering
\resizebox{0.8\columnwidth}{!}{%
\begin{tabular}{@{}lcccc@{}}\toprule
Noise level & 10\% & 20\% & 50\% & 70\% \\ \midrule
CARS$_{\text{full}}$ & 0.992 & 0.994 & 0.989 & 0.981\\
CARS$_{80}$ & 0.999 & 1.00 & 0.994 & 0.982\\
CARS$_{60}$ & 0.996 & 1.00 & 0.992 & 0.970\\
CARS$_{40}$ & 1.00 & 1.00 & 0.980 & 0.945\\
CARS$_{20}$ & 0.977 & 0.973 & 0.937 & 0.903\\
 \bottomrule
\end{tabular} }
\caption{Ratio P@1(p\%)/P@1(0\%) between no-noise performance and noise-corrected performance as the number of classes used during training is decreased
}\label{tab:negatives}
\vspace{-2mm}
\end{table}

\section{Hyperparameters}

\begin{table}[hbt!]
\centering
\resizebox{0.8\columnwidth}{!}{%
\begin{tabular}{@{}lccc@{}}\toprule
Dataset & Noise &  $\lambda$  & $\tau_{+}/\tau_{-}$  \\ \midrule
CUB & Uniform 0\% & 0.01 & GlobalAvg\\
CUB & Uniform 10\% & 0.01 & GlobalAvg\\
CUB & Uniform 20\% & 0.01 & GlobalAvg\\
CUB & Uniform 50\% & 0.1 & GlobalAvg\\
CUB & Uniform 70\% & - & - \\
CARS & Uniform 0\% & 1 & GlobalAvg\\
CARS & Uniform 10\% & 0.1 & GlobalAvg\\
CARS & Uniform 20\% & 0.1 & GlobalAvg\\
CARS & Uniform 50\% & - & - \\
CARS & Uniform 70\% & - & -\\
SOP & Uniform 0\% & 1 & GlobalAvg\\
SOP & Uniform 10\% & 1 & GlobalAvg\\
SOP & Uniform 20\% & 1 & GlobalAvg\\
SOP & Uniform 50\% & 0.01 & GlobalAvg\\
SOP & Uniform 70\% & 0.01 & GlobalAvg\\
CARS-98N & Realistic & 0.01 & ExpAvg \\
Oxford & Realistic & 0.25 & ExpAvg \\
Landmarks & Realistic & 0.05 & GlobalAvg\\
 \bottomrule
\end{tabular} }
\caption{SuperLoss hyperparameters $\lambda$, $\tau_{+}$ and $\tau_{-}$.
}\label{tab:superloss_hyperparameters}
\end{table}

\begin{table}[hbt!]
\centering
\resizebox{0.6\columnwidth}{!}{%
\begin{tabular}{@{}lccc@{}}\toprule
Dataset & Noise & $\tau$  \\ \midrule
CUB & Uniform 0\% & 0.97\\
CUB & Uniform 10\% & 0.81\\
CUB & Uniform 20\% & 0.70\\
CUB & Uniform 50\% & 0.44\\
CUB & Uniform 70\% & 0.31\\
CARS & Uniform 0\% & 1.00\\
CARS & Uniform 10\% & 0.84\\
CARS & Uniform 20\% & 0.74\\
CARS & Uniform 50\% &  0.44\\
CARS & Uniform 70\% & 0.34\\
SOP & Uniform 0\% & 1.00\\
SOP & Uniform 10\% & 1.00\\
SOP & Uniform 20\% & 0.90\\
SOP & Uniform 50\% & 0.64\\
SOP & Uniform 70\% & 0.48\\
CARS-98N & Realistic &  0.40\\
Oxford & Realistic & 0.62 \\
Landmarks & Realistic & 0.66\\
 \bottomrule
\end{tabular} }
\caption{T-SINT hyperparameter $\tau$
}\label{tab:tsint_hyperparameters}
\vspace{-2mm}
\end{table}

\begin{table*}[hbt!]
\centering
\resizebox{\linewidth}{!}{%
\begin{tabular}{@{}lccccc@{}}\toprule
Noise Rate &  0\% & 10\%   & 20\% & 50\% & 70\%  \\ \midrule
Contrastive Margin $\text{Loss}_{\text{BN-inception}}$ \cite{ctrm}  &  57.41/20.87 & 56.65/19.11 & 56.33/18.64 & 40.02/7.97 & 34.45/6.32 \\
Contrastive Margin $\text{Loss}_{\text{ViT-B/32}}$ \cite{ctrm}  &  71.17/29.43 & 67.88/25.99 & 62.49/20.62 & - / - & - / - \\
$\text{PRISM}_{\text{BN-inception}}$ \cite{liu2021noise}  &  57.48/18.86 &  58.32/20.17 & 57.33/19.18 & 54.29/17.25 & 46.78/12.80 \\
$\text{PRISM}_{\text{ViT-B/32}}$ \cite{liu2021noise} &  72.06/31.11 & 72.43/31.27 & 71.93/31.10 & 70.78/29.66 & 64.45/23.85 \\
$\text{SuperLoss}_{\text{ViT-B/32}}$\cite{superloss}  &  70.32/29.05  & 69.85/28.53  & 69.21/27.67 & 58.85/18.71 & - / - \\
$\text{T-SINT}_{\text{ViT-B/32}}$(\text{Ours})      &  72.05/31.60 & 71.73/31.50 & 71.49/31.50 & 71.08/30.37 & 70.51/29.74 \\ \bottomrule
\end{tabular} }
\caption{Precision@1 / MAP@R (\%) on CUB dataset with synthetic uniform label noise. 
}\label{tab:cub_result}
\end{table*}

\begin{table*}[hbt!]
\centering
\resizebox{\linewidth}{!}{%
\begin{tabular}{@{}lccccc@{}}\toprule
Noise Rate &  0\% & 10\%   & 20\% & 50\% & 70\%  \\ \midrule
Contrastive Margin $\text{Loss}_{\text{BN-inception}}$ \cite{ctrm}  &  75.37/21.16 & 74.85/18.75 & 67.27/13.22 & 36.60/3.17 & 32.54/2.62 \\
Contrastive Margin $\text{Loss}_{\text{ViT-B/32}}$ \cite{ctrm}  & 88.85/41.64 & 88.78/41.04 & 87.79/37.88 & - / - & - / - \\
$\text{PRISM}_{\text{BN-inception}}$ \cite{liu2021noise}  & 80.02/22.95 & 78.02/21.37 & 76.93/19.76 & 70.15/15.77 & 52.75/7.50   \\
$\text{PRISM}_{\text{ViT-B/32}}$ \cite{liu2021noise} &  89.08/41.62 & 89.08/41.08 & 88.97/40.81 & 87.44/38.20 & 80.63/28.48 \\
$\text{SuperLoss}_{\text{ViT-B/32}}$\cite{superloss}  &  87.89/39.03 & 87.18/36.94 & 86.69/34.55 & - / - & - / - \\
$\text{T-SINT}_{\text{ViT-B/32}}$(\text{Ours})      &  89.67/42.71 & 88.97/42.21 & 89.10/41.90 & 88.69/40.41 & 87.94/36.33 \\ \bottomrule
\end{tabular} }
\caption{Precision@1 / MAP@R (\%) on CARS dataset with synthetic uniform label noise. 
}\label{tab:cars_result}
\end{table*}

\begin{table*}[hbt!]
\centering
\resizebox{\linewidth}{!}{%
\begin{tabular}{@{}lccccc@{}}\toprule
Noise Rate &  0\% & 10\%   & 20\% & 50\% & 70\%  \\ \midrule
Contrastive Margin $\text{Loss}_{\text{ResNet-50}}$ \cite{ctrm}  & 64.14/35.64 & 65.50/36.73 & 64.70/ 35.51 & 57.87/28.90 & 52.93/25.02 \\
Contrastive Margin $\text{Loss}_{\text{ViT-B/32}}$ \cite{ctrm}  &  77.00/50.67 & 78.74/52.95 & 77.99/51.27 & 73.65/44.30 & 67.35/37.08 \\
$\text{PRISM}_{\text{ResNet-50}}$ \cite{liu2021noise}  & 76.54/48.71 & 74.93/46.27 & 73.65/44.60 & 60.37/30.68 & 52.73/24.87   \\
$\text{PRISM}_{\text{ViT-B/32}}$ \cite{liu2021noise} & 77.87/50.28 & 77.84/49.30 & 74.75/45.67 & 68.18/37.65 & 61.32/31.37 \\
$\text{SuperLoss}_{\text{ViT-B/32}}$\cite{superloss}  &  80.29/56.39 & 82.09/58.91 & 82.06/57.97 & 82.02/57.94 & 77.14/49.05 \\
$\text{T-SINT}_{\text{ViT-B/32}}$(\text{Ours})      & 78.73/53.80 & 80.91/56.78 & 81.09/56.78 & 81.26/56.59 & 79.52/53.70 \\ \bottomrule
\end{tabular} }
\caption{Precision@1 / MAP@R (\%) on SOP dataset with synthetic uniform label noise. 
}\label{tab:sop_result}
\end{table*}

For all methods, we tune the learning rate and batch size based on the datasets with 20\% and 50\% uniform noise. For the contrastive margin loss and PRISM, we find the best learning rate to be 1e-6 and a batch size of 40 for all datasets. For the SuperLoss, we find the best learning rate to be 1e-5 with a batch size of 128, except for CARS-98N, where we set this learning rate to 1e-6. For T-SINT, we use a learning rate of 1e-6 and a batch size of 80, except for SOP, where we use a learning rate of 3e-6. This is for using the ViT-B/32 backbone. For reproducing the experiments from \cite{liu2021noise}, which consists of the Contrastive Margin Loss and PRISM with BN-inception and ResNet-50 backbones, we use the settings as provided in the paper \cite{liu2021noise}.

We tune the hyperparameters related to noise as follows. For PRISM on uniform noise, we set the estimated noise rate R according to the noise rates we use for uniform noise, e.g. R=0.7 for 70\% uniform noise. For CARS-98N, we use R=0.5 from \cite{liu2021noise}. For Oxford and Landmarks we tune the noise rate and find the best values to be R=0.5 for Oxford and R=0.6 for Landmarks. 

For the SuperLoss, we tune $\lambda$, $\tau_{+}$ and $\tau_{-}$. For $\lambda$ we tried the values 0.001, 0.01, 0.05, 0.1, 0.25, 1.0 as these values were recommended in \cite{superloss}. For the thresholds, \cite{superloss} recommends three options: a global average, an exponential running average with a fixed smoothing parameter or a fixed value given by prior knowledge on the task. We experimented with the global average and the exponential running average. The best hyperparameters can be found in Table \ref{tab:superloss_hyperparameters}. Note that for Landmarks, we took the best hyperparameters according to \cite{superloss}.

For T-SINT, we use the estimation of $\tau$ from Equation \ref{eq:ptp} and tune it from there. An overview of these values for each dataset is presented in Table \ref{tab:tsint_hyperparameters}.

\section{Results}

In Tables \ref{tab:cub_result}, \ref{tab:cars_result} and \ref{tab:sop_result}, the results for uniform noise are given that are presented in Figure \ref{fig:uniform_results}. As a reference, we also present the results on all datasets for the Contrastive Margin Loss and PRISM when using the original backbone from \cite{liu2021noise}. Since \cite{liu2021noise} does not report MAP@R scores, we rerun all these experiments with the hyperparameters provided in the original paper.

\end{document}